\documentclass[10pt,journal,compsoc]{IEEEtran}
\ifCLASSOPTIONcompsoc
  \usepackage[nocompress]{cite}
\else
  \usepackage{cite}
\fi
\ifCLASSINFOpdf
\else
\fi
\usepackage{amsmath}
\usepackage{hyperref}       
\usepackage{url}            
\usepackage{booktabs}       
\usepackage{amsfonts}       
\usepackage{nicefrac}       
\usepackage{microtype}      
\usepackage{amsmath}
\usepackage{cleveref}
\usepackage{graphicx}
\usepackage{subfig}
\DeclareMathOperator*{\argmin}{arg\,min}  

\begin{document}
%
\title{Improved Techniques for GAN based Facial Inpainting}
%
%
%
%

\author{Avisek Lahiri*,
        Arnav~Jain*, Divyasri Nadendla
        and~Prabir~Kumar~Biswas,~\IEEEmembership{Senior Member,~IEEE}
\IEEEcompsocitemizethanks{\IEEEcompsocthanksitem * Denotes equal contribution. A.Lahiri and P.K. Biswas are with Dept. of E\&ECE, Indian Institute of Technology, Kharagpur, India. \protect\\
All correspondence to A. Lahiri . E-mail: avisek@ece.iitkgp.ernet.in
\IEEEcompsocthanksitem A.Jain is currently at Microsoft. Work done while graduating from IIT Kharagpur.
\IEEEcompsocthanksitem  D.Nandendla is currently at Qualcomm. Work done while graduating from IIT Kharagpur.}
}

%
%

\markboth{}%
{Shell \MakeLowercase{\textit{et al.}}: Bare Demo of IEEEtran.cls for Computer Society Journals}
%



\IEEEtitleabstractindextext{%
\begin{abstract}
In this paper we present several architectural and optimization recipes for generative adversarial network(GAN) based facial semantic inpainting.  Current benchmark models are susceptible to initial solutions of non-convex optimization criterion of GAN based inpainting. We present an end-to-end trainable parametric network to deterministically start from good initial solutions leading to more photo realistic reconstructions with significant optimization speed up. For the first time, we show how to efficiently extend GAN based single image inpainter models to sequences by a)learning to initialize a temporal window of solutions with a recurrent neural network and b)imposing a temporal smoothness loss(during iterative optimization) to respect the redundancy in temporal dimension of a sequence. We conduct comprehensive empirical evaluations on CelebA images and pseudo sequences followed by real life videos of VidTIMIT dataset. The proposed method significantly outperforms current GAN based state-of-the-art in terms of reconstruction quality with a simultaneous speedup of over 15$\times$. We also show that our proposed model is better in preserving facial identity in a sequence even without explicitly using any face recognition module during training.
\end{abstract}

}

\maketitle

\IEEEdisplaynontitleabstractindextext

%
\IEEEpeerreviewmaketitle

\IEEEraisesectionheading{\section{Introduction}\label{sec:introduction}}

%
%
%
%
\IEEEPARstart{S}{emantic} 
inpainting is a challenging task of recovering large corrupted areas of an object based on higher level image semantics. Classical inpainting methods \cite{adobe2,adobe10,cvpr1,cvpr6,cvpr2} rely on low level cues to find best matching  patches from the uncorrupted sections of the same image. However, such `\textit{copy-paste}' policy works well for background completions(sky, grass, mountains). However, the task of completing a complex object such as human face is far more challenging because the assumption of finding similar appearance patches does not always hold true. A facial image comprises of numerous unique components, which if damaged, cannot be matched with any other facial parts. An alternative is to use external reference datasets \cite{adobe9}. Though this paradigm enables to find similar matching patches, the low level \cite{adobe2} and mid level \cite{adobe10} features of matched patches are not sufficient to infer valid semantics of the missing regions.
\par Recently Yeh \textit{et al.}\cite{yeh2017semantic} leveraged the recent advancement in generative modeling with Generative Adversarial Networks(GAN) \cite{goodfellow2014generative}. Here, a trained neural network, often termed as the `Generator', is trained to generate semantically realistic faces starting from a latent vector drawn from a known prior distribution. \cite{yeh2017semantic} is the current benchmark for semantic inpainting of faces. It outperforms Context Encoders \cite{context_encoders} which was primarily designed for feature learning with inpainting. In this paper, we consider the model of Yeh \textit{et al.} as baseline model and incorporate several architectural and optimization novelties for improving inpainting quality, optimization speed and adapting to inpaint sequences. Our application area is face inpainting. Specifically, our contributions can be summarized as follows:
\begin{itemize}
\item We show, for a single image inpainitng, initializing a GAN based iterative non convex optimization criterion(Eq. \ref{eq_total_loss}) with a learned parametric neural network(Sec. \ref{sec_initialization_image}), results in more photo realistic initial reconstructions(Fig. \ref{fig_good_init}) compared to state-of-the-art GAN based single image inpainter model with random initialization.
\item To our best knowledge this is the first demonstration of extending single image GAN based inpainter for sequences. For this, we design a recurrent neural network architecture(Sec. \ref{sec_lstm}) for jointly initializing solutions for a group of frames. This design choice learns the scene dynamics leading to temporally more consistent initial solutions. 
\item In a sequence, we exploit redundancy of temporal dimension with a smoothness loss(Sec. \ref{sec_smoothness_loss}) which constraints the final joint iterative solutions of a group of neighboring frames to lie close to each other in Euclidean space. The smoothness loss is  not only  better in enforcing temporal consistency(Sec. \ref{sec_temporal_consistency}) but is also apter in preserving the facial identity(Sec. \ref{sec_facial_identity}) of the subject compared to the baseline version.
\item We present comprehensive empirical evaluations on CelebA images and pseudo sequences followed by real life facial videos from VidTIMIT dataset. In all cases, our proposed model outperforms the current benchmark baseline significantly in terms of visual reconstruction quality with an average speedup of over 15$\times$.
\end{itemize}
\section{Background on GANs}
Proposed by Goodfellow \textit{et al.}\cite{goodfellow2014generative}, a GAN model consists of two parametrized deep neural nets, viz., generator, $G_{\theta_G}$, and discriminator, $D_{\theta_D}$. The task of generator is to yield an image, $x\in \mathcal{R}^{H\times W \times 3}$ with a latent vector, $z\in \mathcal{R}^d$, as input. $z$ is sampled from a known distribution, $p_z(z)$. A common choice \cite{goodfellow2014generative} is, $z\sim \mathcal{U}[-1,1]^d$. The discriminator is pitted against the generator to distinguish real samples(sampled from $p_{data}$) from fake/generated samples. Specifically, discriminator and generator play the following game on $V(D_{\theta_D},G_{\theta_G})$: 
$$\underset{G_{\theta_G}}{min}~~ \underset{D_{\theta_D}}{max}~~ V(D_{\theta_D}, G_{\theta_G}) = \mathbb{E}_{x\sim p_{data}(x)}[\log D_{\theta_D}(x)]$$
\begin{equation}
+ ~~\mathbb{E}_{z\sim p_{z}(z)}[1 - D_{\theta_d}(G_{\theta_G}(z))].
\label{eq_gan_main_goodfellow}
\end{equation}
With enough capacity, on convergence, $G_{\theta_G}$ fools $D_{\theta_D}$ at random \cite{goodfellow2014generative}.
\section{Approach}
\subsection{GAN based semantic inpainting}
We begin by reviewing the current state-of-the-art GAN based single image inpainting model of Yeh \textit{et al.} \cite{yeh2017semantic}, which serves as our reference baseline model. Given a damaged image, $I_d$, and a pre-trained GAN model, the idea is to iteratively find the `closest' $z$ vector(starting from $\mathcal{U}[-1,1]^d$) which results in a reconstructed image whose semantics are similar to corrupted image. $z$ is optimized as,
\begin{equation}
\hat{z} = \underset{z}{\mathrm{argmin}}~~ \{L_{con}(z|I_d, M) + \eta L_{per}(z)\}.
    \label{eq_total_loss}
\end{equation}
$L_{con}(\cdot)$ is contextual loss which penalizes mismatch between original and reconstructed images over the non corrupted pixels.
\begin{equation}
    L_{con}(z|I^d,M) = || M \odot G_{\theta_G}(z)~- I_d  ||_1~,
\end{equation}
where $\odot$ is the Hadamard product operator. $M(x, y) =1$ for uncorrupted pixels and 0 otherwise. $\eta$ is a trade off between the two components of the loss.
$L_{per}(\cdot)$ is the perceptual loss and it a measure of realism of the inpainted output. The pre-trained discriminator is leveraged for assigning this realism loss and is defined as,
\begin{equation}
    L_{per}(z) = \log[1-D_{\theta_D}(G_{\theta_G}(z))]
    \label{eq_perceptual_loss}
\end{equation}
Since, $D_{\theta_D}(\cdot)$ gives the probability of being sampled from real images, Eq. \ref{eq_perceptual_loss} drives the solution of Eq. \ref{eq_total_loss} to lie near to natural image manifold. Upon convergence, the inpainted image, $\hat{I}$, is given as, $\hat{I} = M \odot I_d + (1-M) \ \odot G_{\theta_G}(\hat{z})$. Architectures of $G_{\theta_G}$ and $D_{\theta_D}$ are provided in supplemental document.
\begin{figure*}%
    \centering
    \subfloat[]{{\includegraphics[scale=0.5]{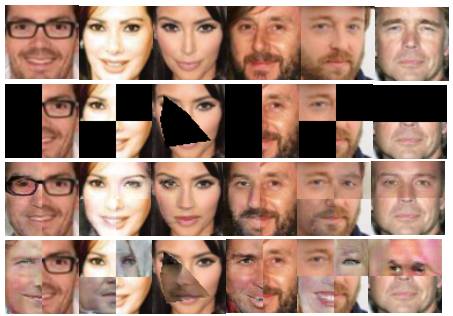} }\label{fig_good_init}}%
    \subfloat[]{{\includegraphics[scale=0.4]{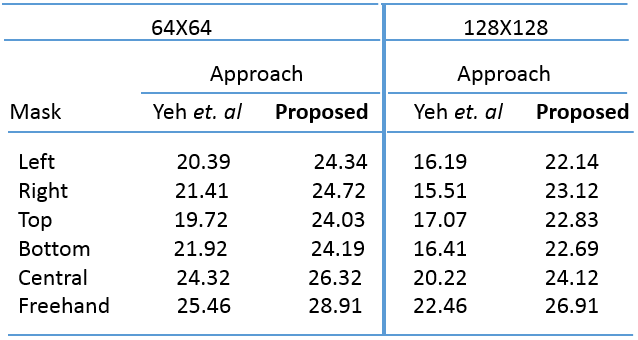} }\label{fig_table_good_init}}%
    \caption{ Benefit of initializing Eq. \ref{eq_total_loss} with proposed learned parametric network, $P_{\theta_z}$  \textbf{(a:)} Visualization of initial solutions of Eq. \ref{eq_total_loss}. Row 1: original images; Row 2: corrupted images; Row 3: Initial solutions using our \textbf{proposed network}, $P_{\theta_z}$; Row 4: Initial solutions using Yeh \textit{et al.}\cite{yeh2017semantic}. Proposed outputs are more photo realistic compared to \cite{yeh2017semantic}.~\textbf{(b:)} Average PSNR after convergence of iterative optimization.  Left, right, top, bottom masks damage the respective 50\% of frame. Central mask damages central 50\% and freehand masks damages approximately 50\% of frame with freehand drawn masks. }%
    \label{fig_benefit_init}%
\end{figure*}
\begin{figure*}[!h]
    \centering
    \includegraphics[scale=0.55]{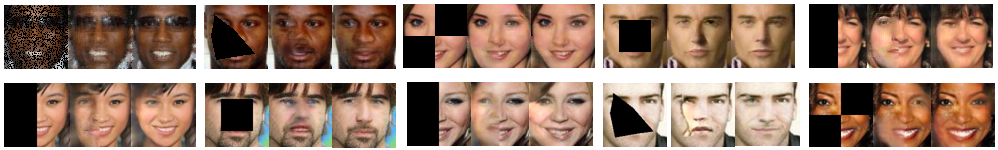}
    \caption{Final inpainted outputs after convergence of Eq. \ref{eq_total_loss}. Top Row: 64$\times$64. Bottom Row: 128$\times$128. For each triplet, Left: masked image, Middle: Inpainting by Yeh \textit{et al.} \cite{yeh2017semantic}, Right: \textbf{Proposed} inpainted output. Proposed outputs are more photo realistic. \cite{yeh2017semantic} specifically suffers at 128$\times$128 resolution. More examples are provided in supplementary document. }
    \label{fig_single_image_final}
\end{figure*}
\section{Single image inpainting}
\subsection{Initializing $z$ vector for single image inpainting} 
\label{sec_initialization_image}
The iterative optimization procedure of Yeh \textit{et. al} \cite{yeh2017semantic} in Eq. \ref{eq_total_loss} yields different results based on the random initialization of $z$; this is mainly attributed to the non convexity of the optimization space. Also, such random initialization results in longer iterations of convergence(Sec. \ref{sec_result_z_init_single_image}) compared to a good initialization of $z$.
The above mentioned problems can be mitigated if we learn to estimate a good $z$ vector directly from damaged image, $I_d$, by feed forward mapping through a deep neural net $P_{\theta_z}$. The parameter set, $\theta_z$, is optimized to minimize some distance metric, $L(\cdot)$: 
\begin{equation}
\theta_z^* = \argmin_{\theta_z} \sum_{i=1}^N L(G_{\theta_G}(P_{\theta_z}(I_d^i)), I_d^i),
\label{eq_z_predictor}
\end{equation}
where $I_d^i$ is the $i^{th}$ corrupted image in dataset. Though Eq. \ref{eq_total_loss} and \ref{eq_z_predictor} are functionally same, prediction using a learned parametric network tends to perform better than ad hoc iterative optimization. This is because, with evolution of training, the network learns to adapt parameters to map images with closely matching appearances to similar $z$ vectors. Parameter update for a given image thus implicitly generalizes to images with similar characteristics. We formulate the loss function, $L(\cdot)$, as, 
\begin{equation}
L(\cdot) = \frac{1}{N}\sum_{i=1}^N ||(I_u^i ~- G_{\theta_G}(P_{\theta_z}(I_d^i))||_2^2 + \lambda \log(1 - D_{\theta_d}(G_{\theta_G}(P_{\theta_z}(I_d^i)))).
\label{eq_z_init_image}
\end{equation} 
The first component of the loss is a mean squared error(MSE) between original and inpainted images. The second component is the same as perceptual loss as defined in Eq. \ref{eq_perceptual_loss}. The MSE loss helps in recovering the global low frequency components of an image while the perceptual loss helps to refine it further with incorporation of detailed high frequency texture components. The parameter, $\lambda$, strikes a balance between the two components of the loss.
\subsection{Extending to series of frames }
\subsubsection{Initialization with a recurrent model}
\label{sec_lstm}
The naive approach of applying the formulation of \cite{yeh2017semantic} on sequences is to inpaint individual frames independently. However, such approach fails to leverage the temporal redundancy inherent in any sequence. 
In this regard, for sequences, we propose to use a Recurrent Neural Network (RNN) to jointly initialize $z$ vectors for an entire group of frames. RNN consist of a hidden state $h_t$ to summarize information observed upto that time step. The hidden state is updated after looking at the previous hidden state and the corrupted image, leading to more consistent reconstructions in terms of appearance. 
	
Since, RNNs suffer from vanishing gradients problem\cite{bengio1994learning} and are unable to capture long dependencies, we use Long Short Term Memory (LSTM) \cite{hochreiter1997long} Networks. LSTMs have produced state-of-the-art results in sequential tasks like machine translation \cite{tell3, tell30} and sequence generation \cite{tell9,Jain_2017_CVPR_Workshops}. 
\begin{figure*}%
    \centering
    \includegraphics[scale=0.35]{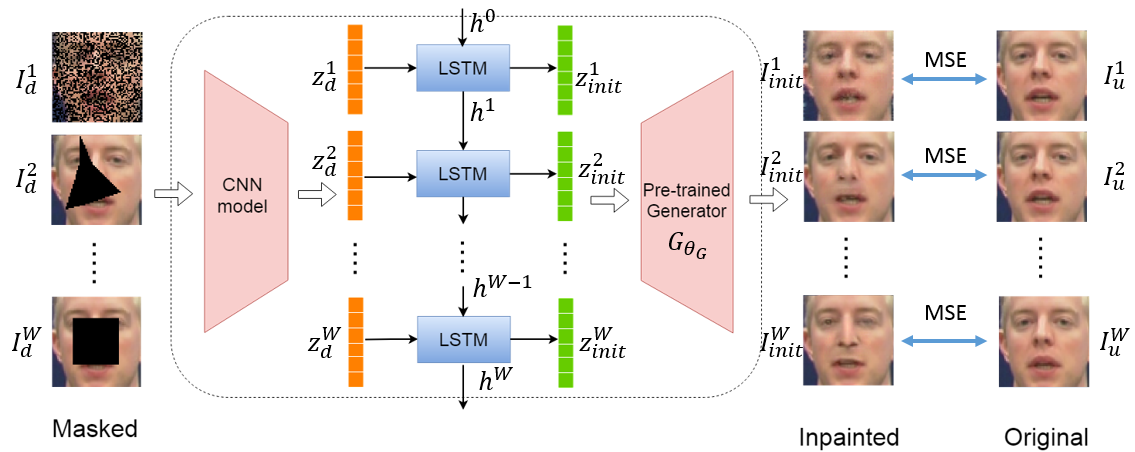}
    \caption{Proposed LSTM based joint initialization of $z$ vectors for a group of frames. See Sec. \ref{sec_lstm} for details of architecture.}
    \label{fig_lstm}
\end{figure*}
In Fig. \ref{fig_lstm}, we show the LSTM based network architecture for initializing a given group of frames. Let, $V=\{I_d^1,I_d^2,...,I_d^W\}$ be a sequence of total $W$ corrupted successive frames. Similar to \cite{vinyals2015show}, each frame is passed through a weight shared CNN descriptor module. Here our CNN's architecture is same as that of $P_{\theta_z}$. Each damaged frame, $I_d^k$ is represented by $z_d^k$. The latent vector $z_d^k$ is passed as input to LSTM module at time step $k$ and the hidden states $h^k$ and cell memory $c^k$ of LSTM are updated. The hidden state $h^k$  is used to obtain the initial $z_{init}^k$ vector which is passed through the pre-trained(and frozen parameters) generator, $G_{\theta_G}$, to output the initial reconstructed image, $I_{init}^k$. 
MSE loss between original image, $I_u^k$, and $I_{init}^k$ is minimized w.r.t all the parameters of LSTM and the CNN descriptor network. Further details are provided in supplemental document.
\subsubsection{Temporal smoothness loss($l_{sm}$)} 
\label{sec_smoothness_loss}
The initialization method using the above mentioned recurrent model ensures that the initial solutions respect the smooth transition of scene dynamics. However, following the initialization, if we independently optimize for each frame, then the final solutions become unconstrained and manifest abrupt changes of facial appearance/expressions. To mitigate this, the idea is to jointly optimize a window of $W$ frames to encourage the final reconstructions to respect the smooth appearance transitions. Disparity between two inpainted images, $\hat{I}^i$ and $\hat{I}^j$ can be approximated by Euclidean distance between their latent vectors ($\hat{z}^i, \hat{z}^j$) \cite{zhu2016generative}. With this approximation, we define
\begin{equation}
l_{sm} = \frac{1}{\binom{W}{2}} \sum_{\forall(i,j)\in W} ||\hat{z}^i - \hat{z}^j||^2.
\label{eq_smoothness_loss}
\end{equation}
It can be seen as a summation of distance loss between all possible pairwise combinations of $z$ vectors of the inpainted frames within the window of $W$ frames. In Sec. \ref{sec_temporal_consistency} we shall see the importance of temporal smoothness loss in yielding a more  consistent set of frames(along temporal dimension) compared to the straight forward per frame application of Yeh \textit{et al.}\cite{yeh2017semantic}.
\section{Experiments}
\subsection{Single image inpainting}
\subsubsection{Dataset}
\label{sec_single_image_dataset}
We evaluate our method on CelebA \cite{celeba} dataset which comprises of 202,599 facial images with coarse alignment. Following the protocol of \cite{yeh2017semantic}, we used 2000 images for testing inpainting performance. Remaining images were used for training GAN. Following face detection, facial bounding boxes are central cropped to 64$\times$64 and 128$\times$128 resolutions.
\begin{figure*}%
    \centering
    \subfloat[]{{\includegraphics[scale=0.25]{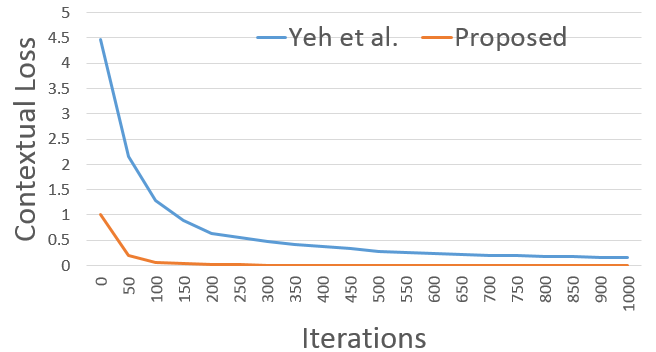} }}%
    \subfloat[]{{\includegraphics[scale=0.25]{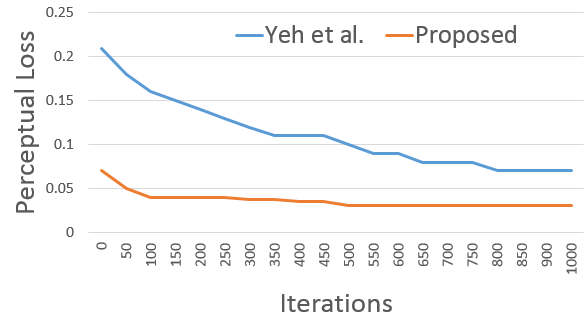} }}%
    \caption{Convergence of \textbf{(a)} contextual loss and \textbf{(b)} perceptual loss of Eq. \ref{eq_total_loss} for a batch of samples.}%
    \label{fig_speedup_single_image}%
\end{figure*}
\subsubsection{Effect of initialization of $z$ vector}
\label{sec_result_z_init_single_image}
In Fig. \ref{fig_good_init} we show the benefit of initializing $z$ vector with a parametrized network as discussed in Sec. \ref{sec_initialization_image}. As evident, a random initialization yields a solution which lies distinctly away from natural face manifold. On the other hand, our parametrized network learns to predict the latent vector by respecting the structural and textural statistics of the uncorrupted pixels. 
One major advantage of good initialization is the speed up of iterative optimization of Eq. \ref{eq_total_loss}. In Fig. \ref{fig_speedup_single_image}, we show an exemplary convergence rates of the two components of Eq. \ref{eq_total_loss}. With the model initialized by our method, both perceptual and contextual losses start at an order less than \cite{yeh2017semantic}. This leads to much faster convergence. In fact, for most of the cases, our proposed model converges after 50 iterations compared to around 700 iterations with \cite{yeh2017semantic}; after this the visual quality does not improve much. Moreover, our solution tends to converge at lower magnitudes of losses and thereby yielding visually more realistic solutions(See Fig. \ref{fig_single_image_final}). This is also evident from the peak signal to noise ratio (PSNR) between original image and the final solution image reported in Fig. \ref{fig_table_good_init}. $p$-value $\leq 10^{-5}$ in all cases. It is encouraging to see  difference of performance is more appreciated at higher resolution of 128$\times$128 resolution.
\subsection{Pseudo sequences}
\label{sec_pseudo_sequences}
\subsubsection{Motivation}
Before directly applying our model on real facial sequences, we dedicate this section to analyze the benefits of our novelties on what we term as, `pseudo sequences'. A pseudo sequence, $S_W$, of length $W$ is formed by taking a single image, $I$, and masking it with $n$ different/same corruptions masks. An ideal inpainter should be agnostic of the corruption masks and yield identical reconstructions for all the $n$ frames. Since  independent optimization of Eq. \ref{eq_total_loss} is unconstrained, there is no explicit restriction on the $\hat{z}$ vectors to be consistent; this is an inherent drawback of GAN based framework of \cite{yeh2017semantic} when applied on sequences.
 \begin{figure*}[!t]
 \centering
 \includegraphics[scale = 0.55]{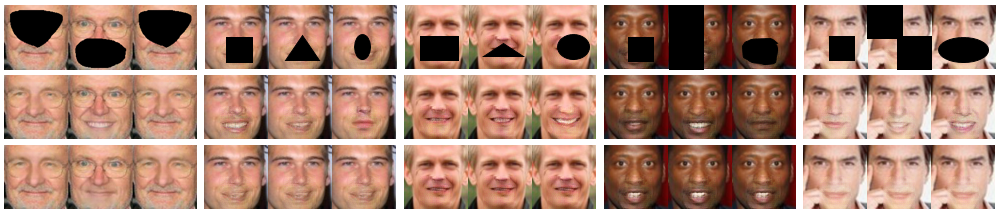}
 \caption{Visualization of consistency of inpainting pseudo sequences. A pseudo sequence is created by masking a given image with different corruption patterns. Ideally we want an inpainter to yield exactly same outputs for a given subject's pseudo sequence.; Top: Masked original pseudo sequence. Middle: Inpainted sequence with Yeh \textit{et al.} \cite{yeh2017semantic}. Bottom: \textbf{Proposed} inpainted sequence. Proposed method yields more consistent sequence w.r.t facial appearances.}
 \label{fig_consistency}
 \end{figure*}
\begin{figure}[!t]
    \centering
    \includegraphics[scale=0.45]{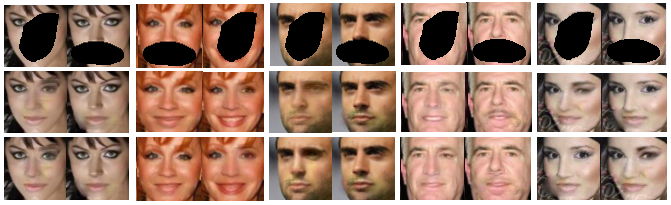}
    \caption{Benefit of LSTM for initialization of sequences. Top Row: A pseudo sequence with same image masked twice differently; Middle Row: Initial solutions by independently predicting $z$ vectors by $P_{\theta_z}$; Bottom Row: Initial solutions by jointly initializing each pair of $z$ vectors with LSTM. Solutions with LSTM are more consistent (similarity near mouth, eye regions).}
    \label{fig_lstm_benefit}
\end{figure}
\begin{table*}[]
\scriptsize
\centering
\caption{Mean consistency (Eq. \ref{eq_consistency}) on CelebA test set measured in terms of PSNR(in dB). A sequence was randomly perturbed by Central, Freehand or Checkboard masks. Higher consistency is better. }
\label{table_consistency}
\begin{tabular}{lc|c|l||c|c|l} \\ \hline \hline
\multicolumn{4}{c}{Resolution @ 64X64}          & \multicolumn{2}{c}{Resolution @ 128X128} &       \\ \hline
     &Central &  Freehand & Checkboard  & Central     & Freehand     & Checkboard  \\ \hline
Yeh \textit{et al.} \cite{yeh2017semantic}  & 22.43         & 22.87           & 20.71  & 22.15 & 20.19 & 19.81 \\
\textbf{Proposed}(Smoothness Loss) & 27.14  & 28.95  & 25.12 & 25.11 & 25.40  & 23.75 \\
\textbf{Proposed}(LSTM init + Smoothness Loss) & 28.01  & 29.15  & 25.73 & 26.01 & 26.10  & 25.09 \\ 
\hline
\label{table_consistency}
\end{tabular}
\end{table*}
\subsubsection{Temporal Consistency}  
\label{sec_temporal_consistency}
We define temporal consistency, $\eta_{temp}$, as the mean pairwise PSNR between all possible pairs($\hat{I}^i, \hat{I}^j$) of inpainted frames within  a pseudo sequence, $S_W$, of length, $W$;
\begin{equation}
\eta_{temp} = \frac{1}{\binom{W}{2}} \sum_{\forall(i,j)\in S_W} PSNR(\hat{I}^i, \hat{I}^j)
\label{eq_consistency}
\end{equation}
Eq. \ref{eq_consistency} allows to enumerate the consistency of a generative model. Ideally we want $\eta_{temp}$=0. Please note that this evaluation is not possible on real videos because the transformation from one frame to another is not known and thus it is not possible to align the frames to a single frame of reference without incorporating interpolation noise with motion compensator\cite{caballero2016real}. In our results, \textit{`Smoothness Loss'} refers to temporal smoothness loss (Eq. \ref{eq_smoothness_loss}). \textit{`LSTM init'} refers to initializing group of 3 frames using proposed LSTM model (Sec. \ref{sec_lstm}).\\
\textbf{Benefit of initialization with LSTM:} First, in Fig. \ref{fig_lstm_benefit} we show the benefit of initializing solutions for a group of pseudo frames with LSTM over per frame independent initialization with  $P_{\theta_z}$. Frames initiated with LSTM tends to be more consistent compared to initiation by $P_{\theta_z}$. This is attributed to the recurrent structure of LSTM which learns that in pseudo sequences, the frame are static. Learning such temporal dynamics is not possible by $P_{\theta_z}$ which is curated for single image initialization. \\
\textbf{Consistency of final solutions:} In Table \ref{table_consistency} we compare the mean temporal consistency on the 2000 pseudo sequences created over the CelebA test set with $W$=3.  The reported mask patterns are: a) Central : randomly corrupt 40\%-70\% of central part of image, b)Checkboard: corrupt 50\% of image with checkboard sizes drawn uniformly from the set of \{8$\times$8, 16$\times$16, 32$\times$32\}, c)Freehand: corrupt around 40\% of pixels with randomly hand drawn masks. Our proposed method with Smoothness Loss (Eq. \ref{eq_smoothness_loss}) fosters in a more consistent sequence of inpainting compared to the vanilla method of per frame model of Yeh \textit{et al.} \cite{yeh2017semantic}. The observations are statistically significant with $p$ value $<$ 0.05 in all cases. Moreover if we initialize the $z$ vectors of the pseudo sequence with a LSTM model, then the consistency of the sequence improves. This can be attributed to more consistent initialization of $z$ vectors by LSTM followed by Smoothness Loss which maintains the similarity of $z$ vectors.
In Fig. \ref{fig_consistency} we visually show the advantage of our proposed modifications. 
One has to appreciate that a set of inpainted frames by \cite{yeh2017semantic} is a mixture of faces with neutral and smiling appearances or different levels of smiles. However, our model yields a set with consistent facial appearance/expressions. In the context of real videos, these observations would mean that there can be drastic change of facial expressions among two adjacent corrupted frames if inpainted by \cite{yeh2017semantic}; such abrupt change of appearance is not common in videos. However, our model has the promise to inpaint a group of neighboring frames with consistency of appearance. Also, if inpainted by \cite{yeh2017semantic}, the stationary portions of frames would tend to show flickering effects due to different hallucinations of textural details independently on each frame.
\begin{figure*}[!t]
    \centering
    \subfloat[]{{\includegraphics[scale=0.35]{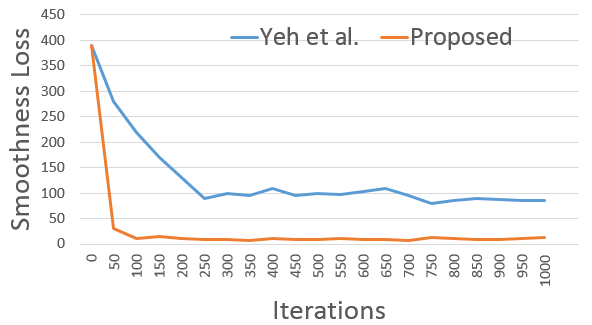} }\label{fig_z_loss}}
    \subfloat[]{{\includegraphics[scale=0.45]{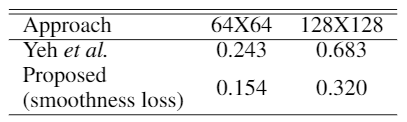}}\label{fig_facenet_loss}}%
    \caption{ (a): Convergence of temporal smoothness loss(Eq. \ref{eq_smoothness_loss}) on a batch of CelebA pseudo sequence. (b): Mean FaceNet loss on CelebA pseudo sequences.}%
    \label{fig_smoothenss_facenet}%
\end{figure*} 
 \begin{figure*}
 \centering
 \includegraphics[scale = 0.5]{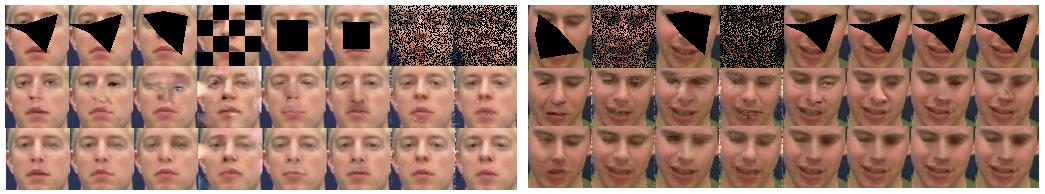}
 \caption{Inpainting on sample snippets of VidTIMIT video dataset. Top Row: Damaged sequence. Middle Row: Inpainting by Yeh. \textit{et al.} \cite{yeh2017semantic}. Bottom Row: \textbf{Proposed method}. It is evident that proposed framework yields visually better samples. More examples are in supplemental document.}
 \label{fig_video}
 \end{figure*}
\subsubsection{Disparity between converged $\hat{z}$ vectors}
To bolster the finding in the above section we also study the disparity of the converged $\hat{z}$ vectors. Ideally, for a given pseudo sequence, the converged $\hat{z}$ vectors should be identical. We can quantify the disparity using the temporal smoothness loss of Eq. \ref{eq_smoothness_loss}. In Fig. \ref{fig_z_loss} we show an exemplary plot of decay of smoothness loss for a pseudo sequence. The proposed method of implicitly minimizing Eq. \ref{eq_smoothness_loss} results in near identical solutions for Eq. \ref{eq_total_loss} over the sequence. However, the converged $\hat{z}$ vectors using \cite{yeh2017semantic} shows more variation. Also, the latter method is slower in convergence. 
\subsubsection{Identity preservation}
\label{sec_facial_identity}
It is important that a sequence of inpainted frames not only appears visually realistic but also maintains the facial identity of the subject. To evaluate this we use FaceNet embeddings \cite{facenet}. FaceNet learns a parametrized network, $F$, to represent a given facial image into a 128-D real vector; $F: I^{H\times W \times 3}~ \rightarrow \mathcal{R}^{128 \times 1}$. Images of same subject yields similar embeddings and is enumerated by the L$_2$ distance between the embeddings. For a given sequence, $S_W$, identity loss, $l_{seq}$, is,
\begin{equation}
l_{seq} =\frac{1}{W} \sum_{\forall i \in S_W} ||F(\hat{I^i}) - F(I_u) ||_2^2.
\label{eq_identity_loss}
\end{equation}
$\hat{I}_i$ is the i$^{th}$ inpainted frame within the pseudo sequence and $I_u$ is the original uncorrupted image. In Fig. \ref{fig_facenet_loss} we report the mean identity loss over the 2000 pseudo sequences. Our proposed method(LSTM init + Smoothness Loss), retains the identity of a person over a sequence more veraciously than \cite{yeh2017semantic}. In our initial experiments, we explicitly included $l_{seq}$ for optimizing Eq. \ref{eq_total_loss}. However, we get similar sequence identity preservation prowess with the Smoothness Loss constraint alone. Authors in \cite{donahue2017semantically}  showed that a $z$ vector can be semantically decomposed to $z_I$, the identity component and $z_o$, the appearance component. Since our proposed Smoothness Loss enforces similarity of converged $\hat{z}$ vectors, the task of identity preservation is implicitly incorporated in the process.
\begin{table*}[!t]
\scriptsize
\centering
\caption{\scriptsize Inpainting PSNR (in dB) on test sequences of VidTIMIT dataset.}
\label{my-label}
\begin{tabular}{lllllllllll}
\hline\hline
Approach                                                                        & \multicolumn{10}{l}{Resolution @ 64X64}                                                 \\\hline
\multicolumn{11}{c}{Subject Name}                                                                                                                                         \\\hline
                                                                                & mrj001 & mwbt0 & mtmr0 & mtas1 & mreb0 & mrgg0 & mdbb0 & mjsw0 & fjre0 & fjas0 \\\hline\\
Yeh \textit{et al.} \cite{yeh2017semantic}                                                                      & 24.32  & 25.32  & 23.61  & 26.11  & 25.12  & 26.01  & 25.98  & 26.09  & 25.31  & 25.81  \\
\begin{tabular}[c]{@{}l@{}}\textbf{Proposed}\\ (Smoothness Loss)\end{tabular}            & 26.12  & 27.01  & 25.11  & 27.11  & 26.91  & 26.98  & 27.11  & 27.21  & 27.00  & 27.71  \\
\begin{tabular}[c]{@{}l@{}}\textbf{Proposed}\\ (LSTM init + Smoothness Loss)\end{tabular} & 27.02  & 28.07  & 27.11  & 28.87  & 28.87  & 28.78  & 29.21  & 29.12  & 28.21  & 29.01 \\\hline\hline\\
                                                                        & \multicolumn{10}{l}{Resolution @ 128X128}                                                 \\\hline
\multicolumn{11}{c}{Subject Name}                                                                                                                                         \\\hline
                                                                                & mrj001 & mwbt0 & mtmr0 & mtas1 & mreb0 & mrgg0 & mdbb0 & mjsw0 & fjre0 & fjas0 \\\hline\\
Yeh \textit{et al.}                                                                      & 22.22  & 23.09  & 21.11  & 23.98  & 23.11  & 24.12  & 23.65  & 25.45  & 24.09  & 23.11  \\
\begin{tabular}[c]{@{}l@{}}\textbf{Proposed}\\ (Smoothness Loss)\end{tabular}            & 24.15  & 25.51  & 23.78  & 25.23  & 25.08  & 25.18  & 25.32  & 25.36  & 25.78  & 25.98  \\
\begin{tabular}[c]{@{}l@{}}\textbf{Proposed}\\ (LSTM init + Smoothness Loss)\end{tabular} & 25.01  & 27.12  & 25.98  & 27.02  & 26.81  & 27.11  & 27.62  & 27.32  & 27.34  & 27.78  \\\hline
\label{table_psnr_video}
\end{tabular}
\end{table*}
\begin{table*}[]
\scriptsize
\centering
\caption{\scriptsize Mean contextual, perceptual and FaceNet losses on VidTIMIT test videos}
\label{my-label}
\begin{tabular}{lccc||cccc}
\hline\hline
Approach                              & \multicolumn{3}{c}{Resolution @ 64X64}           & \multicolumn{3}{c}{Resolution @ 128X128}         \\\hline
                                      & Contextual  & Perceptual  & FaceNet  & Contextual  & Perceptual  & FaceNet  \\
Yeh \textit{et al.} \cite{yeh2017semantic}                           & 0.25            & 0.13            & 0.28         & 0.41            & 0.20            & 0.68         \\
\textbf{Proposed}(LSTM init + Smoothness Loss) & 0.09            & 0.02            & 0.11         & 0.23            & 0.11            & 0.23  \\\hline     \label{table_video_losses}
\end{tabular}
\end{table*}
\subsection{Experiments on VidTIMIT dataset}
The experiments with pseudo sequences taught us two lessons, viz., a)LSTM based group initialization is better than independent initialization with $P_{\theta_z}$ and b) Temporal smoothness loss is essential in imposing temporal consistency. With these understandings we proceed to test our model on real life facial video sequences. To our best knowledge, this is the first attempt towards GAN based inpainting on real videos. For this, we selected the VidTIMIT dataset \cite{face_video}\footnote{Availabe at: \href{http://conradsanderson.id.au/vidtimit/\#examples}{http://conradsanderson.id.au/vidtimit/\#examples}} which consists of video recordings of 43 subjects each narrating 10 different sentences. Images of CelebA dataset are of superior resolution than those of VidTIMIT. Due to this intrinsic difference of data distribution we finetuned our pretrained(trained on CelebA) GAN models on randomly selected 33 subjects of VidTIMIT.  Remaining videos of 10 subjects were kept for testing inpainting performances. In total there are total 9600 frames for testing. We follow the same procedure of Sec. \ref{sec_single_image_dataset} for cropping faces and Sec. \ref{sec_temporal_consistency} for creating random masks. 
In Table \ref{table_psnr_video} we compare PSNR of different inpainting approaches for each subject (each subject has 10 videos). Here again we observe that our proposed models perform superior compared to \cite{yeh2017semantic}. From Table \ref{table_video_losses} it is evident that our proposed model yields more visually realistic solutions(lower perceptual loss) while retaining the appearance of the non corrupted pixels(lower contextual loss) and facial identity(lower FaceNet Loss).
\section{Discussion}
In this paper we proposed several innovations for better optimization of the GAN based inpainting cost function. The study on pseudo sequences enabled us to do ablation studies to appreciate the benefits of each component of our proposals. Since the generator was same for both the comparing models, the improvements are solely due to our contributions. Finally, we bolstered our understandings with experiments on real videos.  However, the performance of inpainting strongly relies on the generative model and the GAN training procedure. An immediate extension would be to improve the generative model itself to generate photo realistic samples at higher resolutions. The recent works of Stacked GAN \cite{stackgan} and progressive stagewise training of GANs \cite{karras2017progressive} show promise towards this end. It would be interesting to integrate the innovations of this paper in such high resolution generative model pipelines.

\ifCLASSOPTIONcompsoc
  \section*{Acknowledgments}
\else
  \section*{Acknowledgment}
\fi
The work is funded by a Google PhD Fellowship to Avisek.

\ifCLASSOPTIONcaptionsoff
  \newpage
\fi

\bibliographystyle{IEEEtran}
\bibliography{IEEEabrv,egbib}
\end{document}